%% file: sn-article.tex
\documentclass[pdflatex,sn-mathphys-num]{sn-jnl}

\usepackage{graphicx}%
\usepackage{multirow}%
\usepackage{amsmath,amssymb,amsfonts}%
\usepackage{amsthm}%
\usepackage{mathrsfs}%
\usepackage[title]{appendix}%
\usepackage{xcolor}%
\usepackage{textcomp}%
\usepackage{manyfoot}%
\usepackage{booktabs}%
\usepackage{algorithm}%
\usepackage{algorithmicx}%
\usepackage{algpseudocode}%
\usepackage{listings}%
\usepackage{hyperref} 
\usepackage{nicefrac}
\usepackage{enumitem}
\usepackage{natbib}
\usepackage{makecell}
\usepackage{graphicx}

\newcommand{\ourmodel}[0]{GSNO}
\newcommand{\ournetwork}[0]{GSNO}
\newcommand{\ourframework}[0]{DGF}

\theoremstyle{thmstyleone}%
\theoremstyle{thmstyletwo}%

\theoremstyle{thmstylethree}%

\begin{document}

\title[Article Title]{Neural Operators for Biomedical Spherical Heterogeneity}



\author[1]{\fnm{Hao} \sur{Tang}}
\author[2]{\fnm{Hao} \sur{Chen}}
\author[3]{\fnm{Hao} \sur{Li}}
\author[1,2]{\fnm{Chao} \sur{Li}}


\affil[1]{\orgname{University of Dundee}, \orgaddress{\country{United Kingdom}}}
\affil[2]{\orgname{University of Cambridge}, \orgaddress{\country{United Kingdom}}}
\affil[3]{\orgname{Fudan University}, \orgaddress{\country{China}}}



\abstract{
Spherical deep learning has been widely applied to a broad range of real-world problems. Existing approaches often face challenges in balancing strong spherical geometric inductive biases with the need to model real-world heterogeneity.
To solve this while retaining spherical geometry, we first introduce a designable Green's function framework (\ourframework) to provide new spherical operator solution strategy: Design systematic Green's functions under rotational group. 
Based on \ourframework, to model biomedical heterogeneity, we propose \textbf{Green's-Function Spherical Neural Operator} (\ourmodel) fusing 3 operator solutions: (1) Equivariant Solution derived from Equivariant Green's Function for symmetry-consistent modeling; (2) Invariant Solution derived from Invariant Green's Function to eliminate nuisance heterogeneity, e.g., consistent background field; (3) Anisotropic Solution derived from Anisotropic Green's Function to model anisotropic systems, especially fibers with preferred direction. Therefore, the resulting model, \ourmodel~can adapt to real-world heterogeneous systems with nuisance variability and anisotropy while retaining spectral efficiency. 
Evaluations on spherical MNIST, Shallow Water Equation, diffusion MRI fiber prediction, cortical parcellation and molecule structure modeling demonstrate the superiority of \ourmodel.


}




\maketitle

\section{Introduction}\label{sec1}
Deep learning on spherical domains has become essential in scientific and engineering disciplines where data naturally resides on curved manifolds. These include applications ranging from planetary-scale simulations in Earth science \cite{rasp2020weatherbench,pathak2022fourcastnet,bonev2023spherical,liu2024evaluation,hu2025spherical}, brain mapping in neuroscience \cite{zhao2019spherical,zhao2021spherical,elaldi2021equivariant,ha2022spharm,elaldi2024equivariant,sedlar2021diffusion,sedlar2021spherical}, omnidirectional perception in computer vision \cite{su2017learning,zhang2018saliency,wu2020spherical,su2021learning,xu2021spherical}, and molecular structure learning in biology and chemistry \cite{boomsma2017spherical,cohen2018spherical,kondor2018clebsch,cobb2020efficient,esteves2023scaling,su2025lumicharge}. 
These applications require models respecting spherical geometry to avoid distortions or coordinate-dependent artifacts, while remaining expressive enough to capture heterogeneous, spatially varying, and often nonlinear phenomena that characterize real-world spherical systems.

A broad range of methods has been developed to meet these demands. Planar‑projection CNNs~\cite{xu2014scale,ronneberger2015u,maron2017convolutional,coors2018spherenet,park2020spheregan} adapt Euclidean models to spherical tasks but inevitably introduce grid distortions and break spherical geometry, limiting performance on sphere‑native tasks~\cite{cohen2016group,cohen2018spherical,bronstein2021geometric,rao2019spherical}. More recent graph-based~\cite{perraudin2019deepsphere,defferrard2020deepsphere,perry2020drawing,lei2020spherical,liu2021spherical} and Transformer-based~\cite{su2019kernel,cho2022spherical,cheng2022spherical,lai2023spherical,carlsson2024heal,zhang2025sgformer,benny2025sphereuformer} spherical models enhance spatial adaptability and nonlinear modeling, yet often embed the sphere as an unstructured graph or sequence, compromising spherical geometric consistency and incurring substantial computational cost~\cite{coors2018spherenet,volkovs2023primed,liu2018deep}. These approaches enhance expressivity but lack a cohernt principle for balancing symmetry and heterogeneity.

Harmonic‑based and group‑equivariant methods address geometric fidelity by enforcing rotational equivariance through spherical convolutions~\cite{cohen2017convolutional,driscoll1994computing,wandelt2001fast,jiang2019spherical}. They underpin spherical CNNs applied to brain cortical analysis~\cite{ha2022spharm,lee2024leveraging,you2024automatic,lee2025spharm}, fiber prediction~\cite{elaldi20243,sedlar2021diffusion,sedlar2021spherical}, as well as spherical Fourier neural operators (SFNOs) for climate and geophysical dynamics~\cite{lin2023spherical, mahesh2024huge1, mahesh2024huge2, zhang2024global}. While such models efficiently capture global interactions and provide strong inductive biases~\cite{edmonds1996angular,muller2006spherical,cohen2016group,cohen2018spherical,tang2025geometric}, the assumption of strict rotational equivariance often fails to model real-world systems that deviate substantially from ideal isotropy and rotational symmetric group SO(3), 
e.g., atmospheric fields with latitude-dependent dynamics and terrain-induced invariance; diffusion MRI showing microstructural anisotropy. Strict equivariance can therefore suppress essential local structure, leading to underfitting and reduced generalization
~\cite{stakgold2011green,lucarini2024detecting, behroozisensitivity,giambagli2021machine,bronstein2021geometric}. 
Attempts to break rotational equivariance via activation functions, linear layers, positional encoding, and hybrid pooling~\cite{wang2022approximately,bonev2023spherical,pertigkiozoglou2024improving,huang2022equivariant,duval2023faenet,zheng2024relaxing,li2024physics} still remain aspheric native, heuristic or task-specific. A principled solution integrating equivariant inductive bias with real-world adaptability remains challenging.

This work aims to address the adaptability required to represent the complex, heterogeneous real-world spherical systems that do not adhere to the symmetry SO(3)-group.
We propose the Green's function Spherical Neural Operators (\ourmodel), fusing Equivariant, Invariant and Anisotropic components to simulate heterogeneous biomedical systems.
We first establish a Designable Green’s Function Framework (\ourframework) to construct a new class of spherical operators.
Based on this framework, we design three Green's functions and derive the corresponding solution operator for different systems: 
(1) Equivariant Green's function $G_E(R^{-1}u)$ based on the relative position {\boldmath $\rightarrow$} Existing Equivariant Operator for isotropic symmetry modeling;
(2) Invariant Green's function $G_I(u)$ based on the absolute position {\boldmath $\rightarrow$} A new Invariant Operator for invariance modeling;
(3) Anisotropic Green's function $G_A(u \cdot Rd)$ based on the learnable directions {\boldmath $\rightarrow$} A new Anisotropic Operator with learnable symmetry groups $SO(\theta)$ for anisotropy modeling.
The three solution operators are concatenated to form~\ourmodel~for efficient heterogeneous modeling.

We evaluate \ourmodel~across diverse tasks in physical, biomedical, and computational domains, including strictly symmetric (molecular QM7 potentials), where equivariance must be preserved; partially symmetric (cortical parcellation), where global structure coexists with localized anisotropy; and 
strongly heterogeneous (weather forecasting, shallow-water simulations and fiber orientation), where symmetry is heavily violated. Across all benchmarks, \ourmodel~consistently outperforms comparison methods. Ablation studies reveal that the corrective term enhances performance specifically in heterogeneous regimes, while maintaining equivariant behaviour when symmetry is present. These results suggest our approach achieves superior generalization, parameter efficiency, and robustness, particularly in heterogeneous or nonlinear regimes. Our main contributions are summarized as:
\begin{enumerate}[leftmargin=*]
\item A designable Green's function framework (\ourframework) to design different operators.
\item Three spherical operators derived from 3 Green's functions, especially the new Invariant and Anisotropic Operators for biomedical heterogeneity.
\item  Propose \ourmodel~by fusing different types of operators for spherical system modeling.
\item Extensive experiments across diverse systems, demonstrating the performance of different operators and \ourmodel.
\end{enumerate}

Together, these contributions provide a rigorous and practically robust foundation for spherical operator learning, aligning geometric fidelity with the complexity of real-world spherical systems while overcoming the fundamental limitations of strictly isotropic and equivariant models.


\section{Results}\label{sec2}

\begin{figure*}
    \centering
    \includegraphics[width=1.0\linewidth]{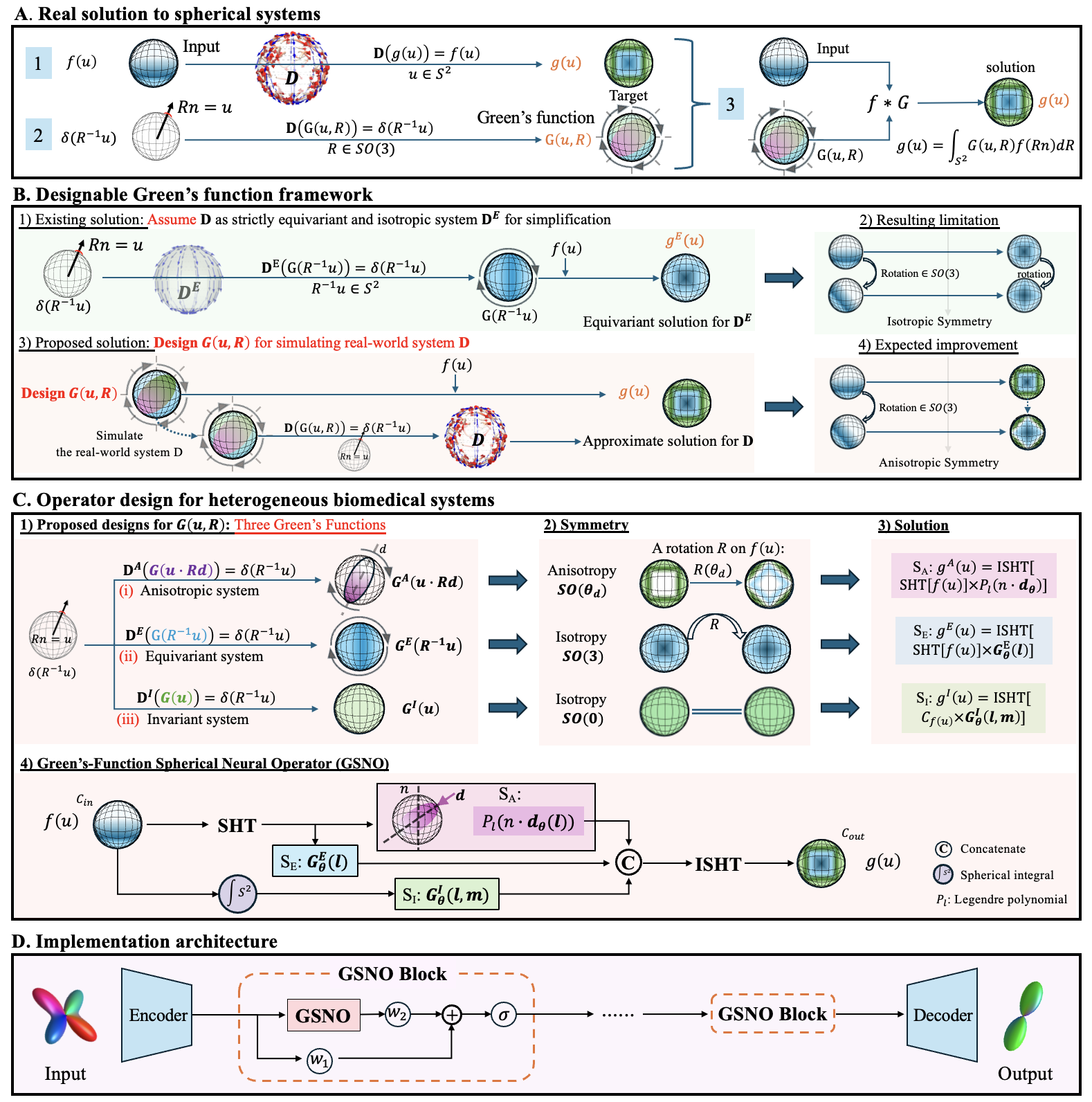}
    \caption{The proposed \ourframework~and \ourmodel. SHT and ISHT represent spherical harmonic transform and inverse transform. $w$ is the linear layer and $\sigma$ is the activation function.}
    \label{fig:isfno}
\end{figure*}

\subsection{Study design}
\subsubsection*{Designable Green's Function Framework:}
Figure 1(A) shows the real solution to a real spherical linear system. Specifically, for a differential operator system defined on the sphere:
\begin{equation}
D(g(u)) = f(u), \quad u \in S^{2},
\label{eq:pde}
\end{equation}
where $f(u)$ is the input function and $g(u)$ is the target. By defining the spherical Green's function $G(u, R)$ of the system~\cite{tang2025generalized}, derive the solution of this system as:
\begin{equation}
g(u) = \int_{S^{2}} G(u, R) f(Rn) \, dR,
\label{eq:solution_origin}
\end{equation}
where \( R \) is an element of the three-dimensional rotation group \( SO(3) \) and n is the north pole. 

Fig.1 (A) shows the real solution to spherical systems based on Green's function method. In the proposed Designable Green's Function Framework (\ourframework) shown in the Fig.1 (B), we demonstrates the underlying limitation of existing simplified solution: Systematic isotropy and equivariance limit the real-world system modeling. To solve this, we propose to design new Green's functions to simulate real-world systems while retaining strictly sphere-native.

\subsubsection*{Equivariant Operator from $G(R^{-1}u)$:}
We design $G(R, u)$ as $G(R^{-1}u)$, and apply the spherical convolution theorem~\cite{driscoll1994computing,cohen2018spherical,bonev2023spherical} to derive the equivariant operator solution with SO(3)-symmetry as:
\begin{equation}
\left\{ G(u, R):G(R^{-1}u) \right\} \rightarrow 
\left\{g^E(u) = \text{ISHT}[G_{\theta}(l) \cdot \text{SHT}[f]_l^m ] \right\},
\label{eq:sc_solution}
\end{equation}
where \(G_{\theta}(l)\) denote the learnable, symmetrical kernel in spectral domain. SHT and ISHT is the spherical harmonic transform and the inverse transform. This design enforces strict isotropy and equivariance to achieve efficient global modeling, but ignores heterogeneity, especially invariance and anisotropy, as shown in Figure 1(B).

\subsubsection*{Invariant Operator from $G(u)$:}

To explicitly model invariant physics, such as fixed terrain, the same MRI protocol, and the invariance required for classification/segmentation tasks, we design the Invariant Green's Function $G^I$ and derive the invariant operator solution:
\begin{align}
\label{Gu_summary}
\left\{ G(R, u):G(u) \right\} \rightarrow 
\left\{ g^I(u) = \text{ISHT}[C_f \cdot G_{\theta}^{2}(l, m)] \right\},
\end{align}
where $C_f$ is the spherical integral of the input function $f(u)$.

\subsubsection*{Anisotropic Operator from $G(u \cdot R\mathbf{d})$:}
To explicitly model anisotropy, especially the local directionality existing at the boundaries of cortical partitions and the inherent anisotropy of nerve fibers, we design the Anisotropic Green's Function $G^A$ and derive the anisotropic operator solution:
\begin{align}
\label{Gu_summary}
\left\{ G(R, u):G(u \cdot (R\mathbf{d})) \right\} \rightarrow 
\left\{ g^A(u) = \text{ISHT}[P_l(n\cdot \mathbf{d_\theta}) \cdot \text{SHT}[f]_l^m ] \right\},
\end{align}
where $P_L$ is the $L$-th order Legendre polynomial, $\mathbf{d_\theta}$ is the learnable orientations.

\subsubsection*{Green's-Function Spherical Neural Operator:}
As shown in the Figure 1 (C and D), we design Green's-Function Spherical Neural Operator (\ourmodel), a new spectral learning fusing Equivariant, Invariant and Anisotropic Operators by concatenation to model biomedically heterogeneous systems while retaining spherical geometry.


\subsubsection{Experimental design}
We compare \ourmodel~to other state-of-the-art methods for 5 spherical experimental scenarios with distinct system characteristics, involving various spherical sampling methods, resolutions, and application fields, to comprehensively evaluate the generalization of \ourmodel. For example, brain nerve fiber modeling and cortical parcellation emphasize \ourmodel's ability to model biomedical heterogeneity, including invariance and anisotropy. And the different time scales and spatial scales in shallow water equation highlight the nonlinear and asymmetric modeling capability and stability of \ourmodel. Spherical digital recognition and atomic structure analysis demonstrate the generalization ability of \ourmodel~for spherical symmetry systems.
All experiments are implemented through PyTorch on 32GB V100 GPUs.

\subsection{Synthetic Spherical Scenarios}
We first validate \ourmodel~on two synthetic spherical scenarios with different scales.
\subsubsection{Spherical MNIST}
\label{mnist}

Spherical MNIST is a simulated dataset of handwritten digits on the sphere~\cite{cohen2018spherical}. The native 28×28 images are projected onto 32×32, 64×64, 128×128 and 256×256 equirectangular spherical grids through backward ray-casting and bilinear interpolation, yielding a rotation-free spherical MNIST benchmark at the four target resolutions while preserving the original 60 k / 10 k train-test split. Under the fixed view of spherical MNIST, \ourmodel~achieves more accurate digital classification.

\input{Tables/mnist_equ}

\subsubsection{Shallow Water Equation}
\label{SWE}
Spherical Shallow Water Equations (SWE) form a nonlinear hyperbolic PDE system that models the motion of thin-layer fluids on a rotating sphere. The core underlying assumption is the shallow water approximation, where the vertical scale of the fluid layer is much smaller than the horizontal scale~\citep{bonev2018discontinuous}. Following \citep {bonev2018discontinuous, bonev2023spherical}, we generate the SWE simulation using a classical spectral solver~\citep{giraldo2001spectral}. Four SWE datasets are generated with 4 spatial resolution of $32\times64,64\times128,128\times256,256\times512$; time step of 150 s; 3 channel dimensions: geopotential height (H), vorticity (V), and divergence (D). All the models use the identical dataset and setting: 50 epochs containing 256 samples each, batch size of 4, Adam optimizer with learning rate of $2\times10^{-3}$, and spherical weighted mean relative loss. Models are tested on the dataset generated from 50 initial conditions (50 samples) and evaluated using MRE for each variable. The results in the Table~\ref{tab:SWE_MRE_all} show that \ourmodel~achieves better performance on all variables and time scales.

\input{Tables/SWE_results}

\subsection{Real-world Spherical Scenarios}
We further validate the generalization of \ourmodel~on multiple real-world heterogeneous systems.
\label{Neuroscience}
\subsubsection{Diffusion MRI-based fiber estimation} 

\begin{figure*}
    \centering
    \includegraphics[width=1.0\linewidth]{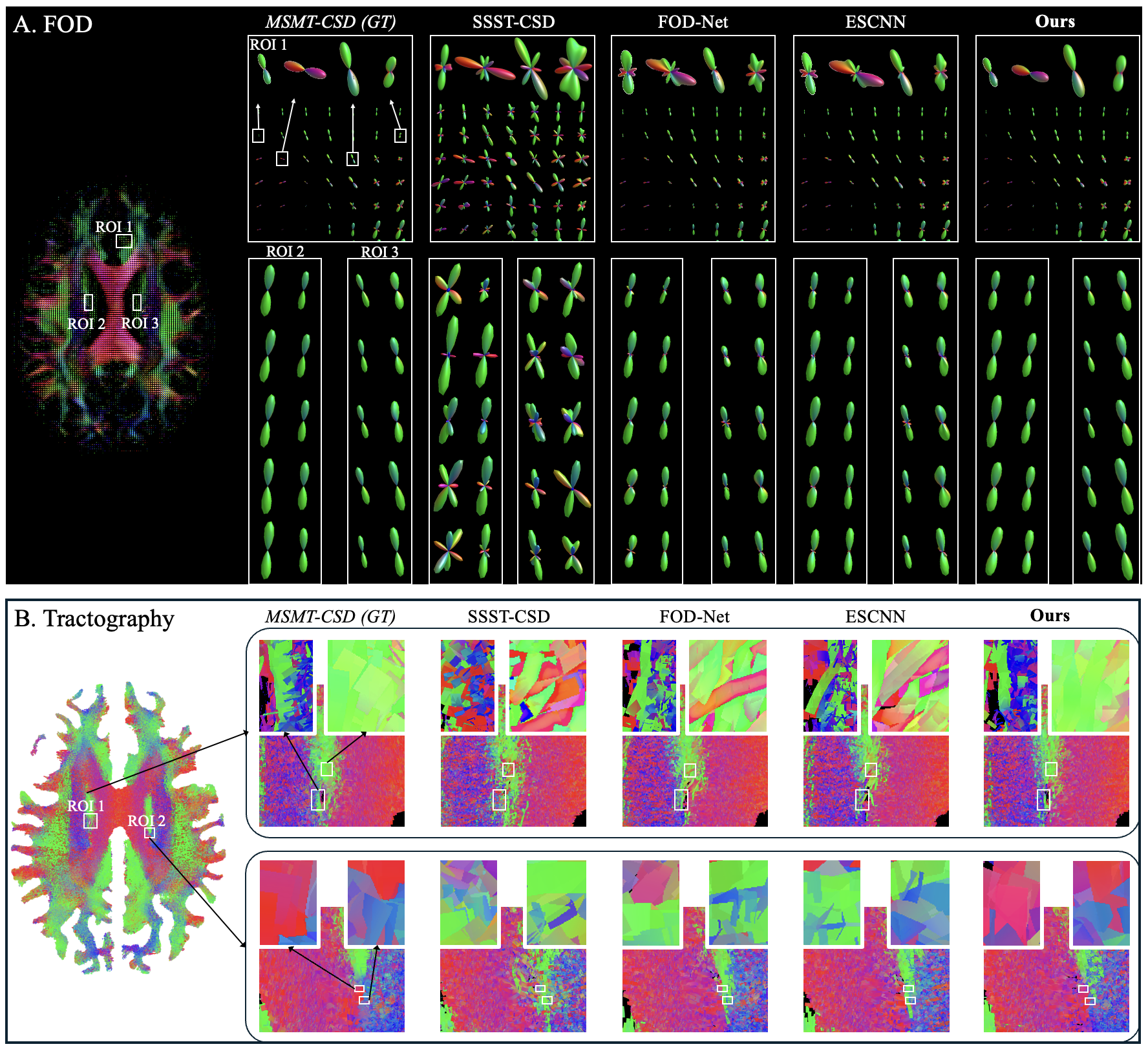}
    \caption{FOD and Tractography on hcp dataset.}
    \label{fig:FOD_tract}
\end{figure*}

\input{Tables/FOD}

Diffusion MRI (dMRI) modeling of brain microstructure is a distinct challenge in: (1) the input signals are sparse and anisotropic diffusion measurements acquired over spherical shells, exhibiting sharp angular variations corresponding to underlying fiber tract orientations; (2) the spatial sampling of dMRI-based fiber orientation distribution is performed using HEALPix, resulting in nonuniform and incomplete coverage of the spherical domain. 

We randomly select dMRI images of 30 subjects in the Human Connectome Project (HCP)~\citep{van2013wu}, where 20 subjects were for training, 5 for validation, and 5 for test. The complete dMRI data of each subject contains 288 volumes (multi-shell HARDI), including 270 volumes of b=1000, 2000, 3000 $s/mm^{2}$ with 90 gradient directions for each shell and 18 b0 volumes. Due to the wide application of low b-value with 32 gradient directions in clinical practice~\citep{zeng2022fod}, we subsample 32 volumes of b=1000 and 1 volume of b0 from the complete dMRI according to the HCP protocol, to obtain single-shell LARDI. Further, we use MSMT-CSD~\citep{jeurissen2014multi} on the multi-shell HARDI to obtain high angular resolution fiber orientation distribution (FOD) as the ground truth (HAR-FOD). Meanwhile, we use SSMT-CSD~\citep{khan2020three} on the single-shell LARDI to obtain single-shell low angular resolution FOD as the condition of \ourmodel~(LAR-FOD). $l_{max}$ is set to the default value of 8 to balance precision and complexity~\citep{zeng2022fod}.

SSMT-CSD~\citep{khan2020three}, FOD-Net~\citep{zeng2022fod}, ESCNN~\citep{snoussi2025equivariant} and our proposed \ourmodel~adopt the same network architecture as FOD-Net. The difference lies in that FOD-Net uses 3D convolutional layers to handle a large number of harmonic coefficients, while ESCNN and \ourmodel~stack voxels on the channel dimension and then use the spherical operator for learning. To ensure a fair comparison, these methods have exactly the same hyperparameters, with the only difference being the types of operators. The results of Angular Correlation Coefficient (ACC)~\citep{zeng2022fod} in Table~\ref{table_dmri} show that \ourmodel~consistently outperforms other models, demonstrating strong generalization under irregular sampling and data sparsity. Especially, Figure~\ref{fig:FOD_tract} shows some regions of interest taken from the caudate nucleus and cingulate gyrus. Compared with the obviously false positive predictions by other methods, the proposed \ourmodel~achieves more precise prediction of fiber orientation and density, especially for the inhibition of false fibers.

\subsubsection{Cortical parcellation} 
\label{CC}

\begin{figure*}
    \centering
    \includegraphics[width=1.0\linewidth]{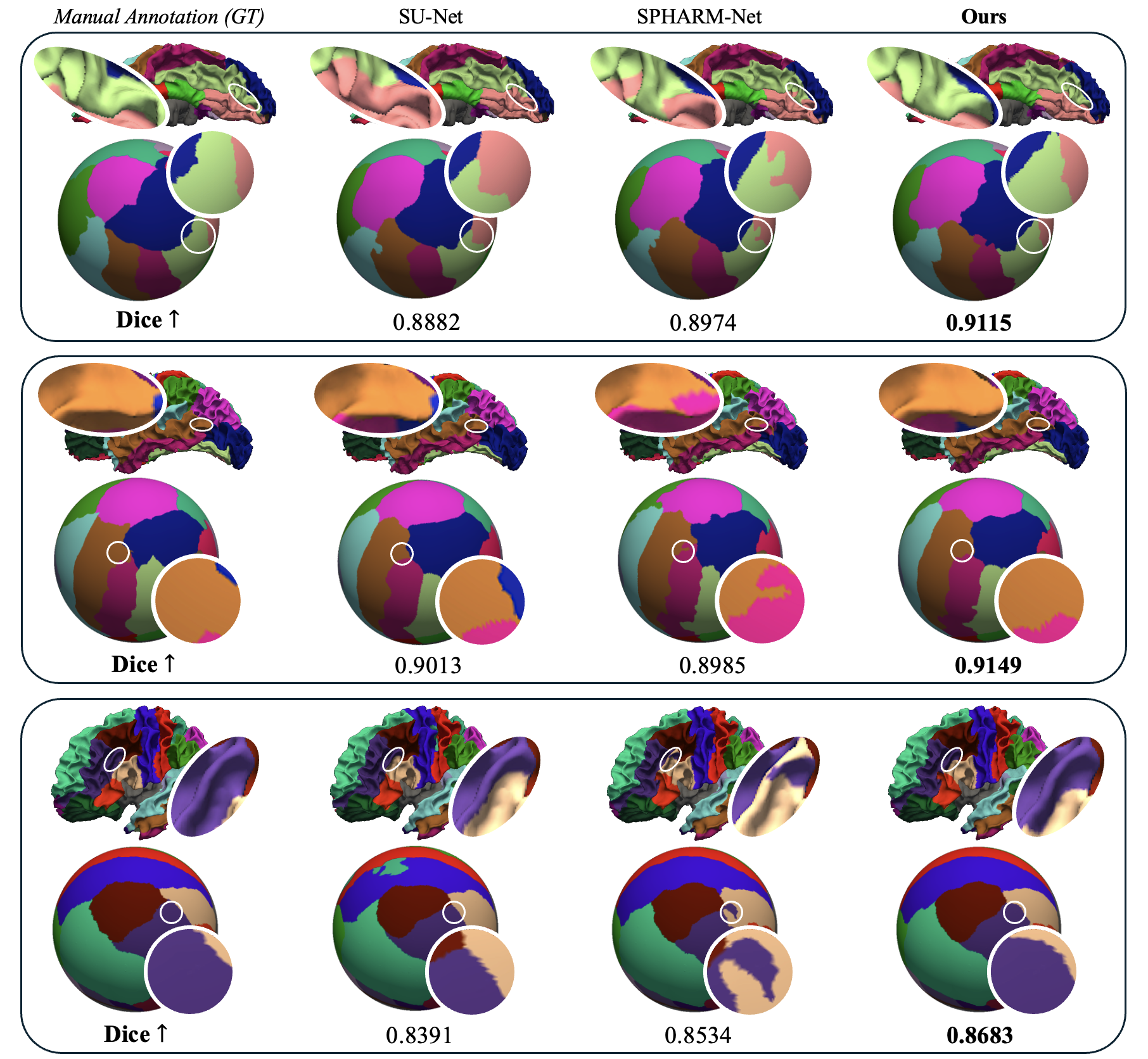}
    \caption{Cortical parcellation on Mindboggle-101 and NAMIC-39 datasets.}
    \label{fig:Cortex}
\end{figure*}

Cortical parcellation requires both equivariance to handle large-scale spatial variations among individuals and biomedical heterogeneity modeling to capture nuisance variability and the local subtle features that determine the boundaries. This study evaluate the \ourmodel's robustness to anatomical variability and its parcellation accuracy on the public Mindboggle-101~\cite{klein2012101} and NAMIC-39~\cite{desikan2006automated} datasets. The datasets comprises cortical surfaces with expert manual annotations, from which we extracted geometric features and mapped them onto the sphere following a preprocessing pipeline consistent with SPHARM-Net~\cite{desikan2006automated, ha2022spharm}. 

All the models are trained and tested using five repetitions of 9-fold cross-validation, with performance assessed by the Dice coefficient. Results demonstrate that \ourmodel~outperforms SOTAs in parcellation accuracy with fewer learnable parameters, without relying on rigid alignment or data augmentation. As shown in the Figure~\ref{fig:Cortex}, \ourmodel~achieves more accurate predictions on some small regions with complex geometers, blurred boundaries or large anatomical variations. It is precisely because the discriminative features of these regions rely on asymmetric local contexts that our flexible heterogeneity modeling can capture them.

\input{Tables/CC_mind101}
\input{Tables/CC_NAMIC39}

\subsubsection{Molecule structure modeling}
\label{Molecules}
\input{Tables/molecule}

We further evaluate the generalization of \ourmodel~on the QM7 benchmark, a energy-prediction dataset~\cite{blum2009970,rupp2012fast} contains 7,165 molecules selected from GDB-13 chemical universe and optimized at the PBE0/def2-TZVP DFT level, providing 0 K atomization energies for neutral systems with up to 23 atoms of H, C, N, O and S. Each molecule is presented by placing non-overlapping spheres around atoms and computing 5-channel Coulomb-type potential maps on the sphere (bandwidth = 10, Driscoll-Healy grid), yielding spherical geometry of molecule~\cite{cohen2018spherical}. 

In this structurally heterogeneous system, subtle variations in local chemical environments and geometric configurations can induce large changes in global molecular energy. Moreover, molecular energies are invariant under rotations, requiring rotational symmetry. Training and testing follow the 5-fold cross-validation split consistent with spherical CNN~\cite{cohen2018spherical}; Root mean squared error (RMSE) is used as evaluation metric. As shown in the Table~\ref{table_qm7}, \ourmodel~with lower parameters achieves more precise energy prediction, which demonstrates the powerful adaptive ability of \ourmodel~in the molecule system with structurally heterogeneity and rotational symmetry.

\section{Discussion}\label{sec12}

These extensive experiments verify the superiority of \ourmodel, which further the feasibility of the designable Green's function framework, especially in simulating real-world heterogeneous systems. Specifically, the derived Invariant Solution is effective to eliminate nuisance heterogeneity, e.g., medical imaging protocol-related noise or anatomical landmarks. The derived Anisotropic Solution is efficient to model anisotropy, especially the white matter fibers with preferred directions.


\section{Methods}\label{sec11}

\subsection{Designable Green's Function Framework}
\label{DGF}
The Green's function method offers a classical strategy to solve PDEs, where the solution is expressed as a convolution integral with the Green's function as the kernel~\citep{li2020neural}. For the sphere-native solution, we define \( D \), a linear differential operator on the sphere, and consider the following PDE:
\begin{equation}
D(g(u)) = f(u), \quad u \in S^{2}
\label{eq:pde}
\end{equation}
where $f(u)$ is the input function and $g(u)$ is the target solution. The spherical Green's function \( G \), associated with \( D \), is defined by the property:
\begin{equation}
D(G(u,R)) = \delta(R^{-1}u) = 
\begin{cases} 
\infty, & Rn = u, \\
0, & Rn \neq u,
\end{cases}
\label{eq:green_prop}
\end{equation}
where \( \delta(.) \) denoting the Dirac delta function defined on the sphere, and $R \in SO(3)$ represents a rotation from the north pole 
\( n \). Using the Green function, the solution to Equation~\ref{eq:pde} is:
\begin{equation}
g(u) = \int_{S^{2}} G(u, R) f(Rn) \, dR
\label{eq:solution_origin}
\end{equation}


This solution is verified in our prior work~\cite{tang2025generalized}. By assuming any system/mapping as the solution of a differential equation: $f(u) \rightarrow g(u)$, we propose the \textbf{Designable Green's Function Framework} (\ourframework) to design different Green's functions to simulate the real-world systems and simplify Equation~\ref{eq:solution_origin} to derive the corresponding operator solution $g(u)$.

\subsection{Three Operators from three Green's Functions}
\label{3C_3G}
We design Equivariant Green's Function $G(R^{-1}u)$, 
Invariant Green's Function $G(u)$ and Anisotropic Green's Function $G(u \cdot R\mathbf{d})$ and derive their operator solutions.

\subsubsection{Equivariant Operator from $G(R^{-1}u)$}
\label{EC_GE}
Inspired by existing equivariant methods, we design \( G(u,R) \) as \( G^E(R^{-1}u) \) assuming strictly isotropic SO(3)-symmetry to achieve strict rotational equivariance. Under this formulation, the prediction target \( g(u) \) is given by:
\begin{align}
\label{eq:sfno}
g(u) &= \int_{S^{2}} G^E(R^{-1}u) f(Rn) \, dR.
\end{align}

Applying the spherical convolution theorem~\cite{driscoll1994computing,cohen2018spherical,bonev2023spherical},  the spherical harmonics transform of \( g(u) \) is derived as:
\begin{align}
\text{SHT}[g(u)](l, m) 
&= \text{SHT}[(f * G^E)](l, m) \nonumber\\
&= {2\pi \sqrt{\frac{4\pi}{2l+1}}}\cdot \text{SHT}[G^E](l, 0) \cdot \text{SHT}[f](l, m).
\end{align}

Thus, the operator solution $g(u)$ is reconstructed via the \text{ISHT}:
\begin{equation}
g(u) = \text{ISHT}(G^E_{\theta}(l) \cdot \text{SHT}[f](l, m)),
\label{eq:solution}
\end{equation}
where \(G_{\theta}(l)\) denote the learnable spectral weights parameterized by the neural operator.

\subsubsection{Invariant Operator from $G(u)$}
\label{GSL}
\textbf{Motivation:}
Spherical biomedical systems—such as cortical surface measurements, diffusion-derived features, or functional activation maps—are often subject to substantial variability arising from acquisition protocols, coordinate choices, and inter-subject alignment procedures. These sources of variability induce differences in the observed data that do not reflect underlying biomedical differences. For instance, global rotations of the sphere, scanner-dependent orientation conventions, or arbitrary surface parameterizations can significantly alter the numerical representation of a sample while leaving its biomedical interpretation unchanged.
If left unconstrained, learning-based models may allocate substantial representational capacity to capturing such variability, thereby conflating coordinate-dependent artefacts with meaningful biomedical signals. This leads to reduced generalization, poor reproducibility across datasets, and sensitivity to irrelevant transformations. From a scientific standpoint, these effects are undesirable, as they obscure the relationship between the learned representation and the biomedical system of interest.

To explicitly address this issue and model invariant physics, inspired by our prior work~\cite{tang2025generalized}, we design the Invariant Green's Function $G^I$:
\begin{align}
\label{G1u}
G(u,R) = G^I(u)
\end{align}

Based on $G^I$, we derive the invariant operator solution under \ourframework:
\begin{align}
\label{G1u_solution}
\text{SHT}[g(u)](l, m) 
&= \int_{SO(3)} f(Rn) \left( \int_{S^2} \left[ \sum_{l',m'} \text{SHT}[G^I]_{l'}^{m'} Y_{l'}^{m'}(u)\right] \overline{Y_l^m(u)} \, du \right) dR \nonumber\\
&= \text{SHT}[G^I](l, m) \int_{SO(3)} f(Rn) dR \nonumber\\
&= C_f \cdot G_{\theta}^{I}(l, m),
\end{align}
where $C_f$ is the spherical integral of the input function ($f(u)$). $G_{\theta}^{I}(l, m)$ is an asymmetric kernel. The invariant operators enforce that the model’s output or intermediate representations remain unchanged under a prescribed set of transformations, such as global rotations of the sphere. Functionally, this amounts to collapsing all observations related by these transformations into a single equivalence class in representation space.

Rather than attempting to learn invariance from data, the invariant operators impose it by construction. This guarantees that nuisance variability is systematically removed from the learned features, preventing the model from encoding information that is irrelevant to the task. As a result, representational capacity is redirected toward modelling biomedically meaningful variation, improving robustness, interpretability, and cross-dataset generalization.
In the context of spherical biomedical data, invariant operators thus serve as a principled mechanism for disentangling biomedical signal from acquisition- and coordinate-induced artefacts, ensuring that other learning is driven by other scientifically relevant information.

\subsubsection{Anisotropic Operator from $G(u \cdot R\mathbf{d})$}
\label{3}

\textbf{Motivation:}
Neural tissue, cortical folding patterns, and white-matter fiber architectures exhibit pronounced directional structure that plays a fundamental role in their function and organization. These anisotropic systems with preferred directions exhibit different properties along different directions. Such anisotropy reflects direct heterogeneity of underlying biomedical mechanisms.
Modelling these systems with operators that assume isotropic SO(3)-symmetry can impose overly restrictive symmetry assumptions, forcing the model to treat all rotations as equally valid transformations.
In many real-world systems, however, directional preferences, local reference frames, or constrained rotational freedoms are intrinsic to the underlying physical or semantic structure. Enforcing full equivariance in such cases can suppress direction-dependent responses, reduce representational flexibility, and lead to the loss of critical anisotropic information.
Anisotropic operators are motivated by the need to model systems whose symmetries are inherently partial, directional, or locally constrained, rather than globally isotropic while retaining sphere-native. They provide a controlled and interpretable way to incorporate biomedically meaningful anisotropy into the model.

\textbf{Modeling Axisymmetric Anisotropic Systems:}
To explicitly model anisotropy, we draw inspiration from structures commonly found in nature that possess a single dominant orientation. A prime example is the 'white matter tract' in the brain, which consists of bundles of parallel neural fibers, akin to bundles of aligned cables. These tracts allow neural signals to propagate rapidly along the fiber orientation but hinder transverse propagation, thus establishing a clear 'preferred orientation'. Inspired by this, we propose a mathematical model, the anisotropic Green’s function whose value depends solely on the angle between a spatial point \(\mathbf{u}\) and a rotatable 'preferred axis' \(R\mathbf{d}\). Its mathematical form and harmonic extension:
\begin{align}
\label{G_ai1}
G^A(u, R) = G^A(u \cdot (R\mathbf{d}))
&= \sum_{L \ge 0} g_L P_L(u \cdot (R\mathbf{d})),
\end{align}
where, \(P_L\) is the \(L\)-th order Legendre polynomial, and \(g_L\) are coefficients that determine the specific shape of the function. \(R\) represents a rotation applied to the system input. \(G_1\) describes a system whose response resembles 'stacked thin sheets': it varies along the \(R\mathbf{d}\) axis but remains completely uniform across the entire plane perpendicular to that axis. This perfectly captures anisotropy with rotational symmetry around a single axis.

\textbf{Generalization to Broad Anisotropic Systems:}
To model more general anisotropy, we further allow different learnable preferred orientations \(\mathbf{d}_L\) for each spherical harmonic degree \(L\). This yields a generalized anisotropic Green's function:
\begin{equation}
\label{G_ai2}
G^A(u \cdot (R\mathbf{d})) \rightarrow \sum_{L \ge 0} g_L P_L(\mathbf{u} \cdot (R\mathbf{d}_L)),
\end{equation}
where each \(L\)-th order component is symmetric about its own axis \(R\mathbf{d}_L\). The generalized form reduces to the axisymmetric form Equation~\ref{G_ai1} when all preferred directions are aligned, i.e., \(\mathbf{d}_L = \mathbf{d}\) for all \(L\).

\textbf{Symmetry Analysis:}
The systems defined by $G^A$ embody a special design that intertwines equivariance and invariance, which is precisely the mathematical origin of their ability to characterize "anisotropy". Specifically, the systems are invariant under rotations around their intrinsic (or per-degree) preferred axis \(\mathbf{d}\) (or \(\mathbf{d}_L\)), as such rotations do not alter the relative relationship between \(\mathbf{u}\) and \(R\mathbf{d}\). However, they are equivariant under other rotation not centered on that preferred axis—a rotation \(R\) of the input source induces a coordinated rotation of the system’s entire response pattern. This orientation-preferred symmetry reflects the intrinsic nature of physical anisotropy. Therefore, by endowing different frequency (angular momentum) components with independent directional degrees of freedom, $G^A$ achieves learnable $SO(\theta)$-symmetry, providing a powerful and theoretically grounded approach for modeling complex anisotropic responses.

We substitute this expansion into the integrand and define the inner integral \(I_{\mathrm{inner}}(R)\) by the Legendre expansion:
\begin{align}
I 
&= \int_{SO(3)} f(Rn)\left(\int_{S^2}\left[ 
G(\mathbf{u} \cdot (R\mathbf{d})
\big)\right] \overline{Y_\ell^m(u)}\,du\right)dR \nonumber\\
&= \int_{SO(3)} f(Rn)\left(\int_{S^2} \Big[\sum_{L\ge0} g_L P_L\big(u\cdot(Rd_L)\big)\Big] \overline{Y_\ell^m(u)}\,du\right)dR \nonumber\\
&= \int_{SO(3)} f(Rn)\,I_{\mathrm{inner}}(R)\,dR.
\end{align}

We solve the inner integral by applying the spherical harmonic addition theorem to each Legendre term:
\begin{align}
I_{\mathrm{inner}}(R) 
&= \int_{S^2}\Big[\sum_{L\ge0} g_L P_L\big(u\cdot(Rd_L)\big)\Big]\overline{Y_\ell^m(u)}\,du \nonumber\\
&= \sum_{L\ge0} g_L \int_{S^2} P_L\big(u\cdot(Rd_L)\big)\,\overline{Y_\ell^m(u)}\,du \nonumber\\
&= \sum_{L\ge0} g_L \frac{4\pi}{2L+1}\sum_{m'=-L}^{L} \overline{Y_L^{m'}(Rd_L)}\int_{S^2} Y_L^{m'}(u)\,\overline{Y_\ell^m(u)}\,du \nonumber\\
&= \sum_{L\ge0} g_L \frac{4\pi}{2L+1}\sum_{m'=-L}^{L} \overline{Y_L^{m'}(Rd_L)} \delta_{L\ell}\,\delta_{m' m} \nonumber\\
&= g_\ell \frac{4\pi}{2\ell+1}\,\overline{Y_\ell^{m}(Rd_l)}.
\end{align}

Based on the transformation law of spherical harmonics under rotations together with the orthogonality of Wigner \(D\)-matrices over \(SO(3)\) equipped with the normalized Haar measure, we substitute the inner result back into the outer integral over \(SO(3)\) and solve the final integral $I$ by the harmonic expansion:
\begin{align}
I
&= g_\ell\frac{4\pi}{2\ell+1} \cdot \int_{SO(3)} f(Rn)\,\overline{Y_\ell^{m}(Rd_l)}\,dR \nonumber\\
&= g_\ell\frac{4\pi}{2\ell+1} \cdot \int_{SO(3)} f(R^{-1}n)\,\overline{Y_\ell^{m}(R^{-1}d_l)}\,dR \nonumber\\
&= g_\ell\frac{4\pi}{2\ell+1} \cdot \int_{SO(3)} \left(\sum_{L,M} \text{SHT}[f]_L^M \sum_{a=-L}^{L} D^{(L)}_{aM}(R)\,Y_L^a(n)\right)\left(\sum_{b=-\ell}^{\ell} \overline{D^{(\ell)}_{b m}(R)}\,\overline{Y_\ell^b(d_l)}\right) dR \nonumber\\
&= g_\ell\frac{4\pi}{2\ell+1} \cdot \sum_{L,M}\sum_{a=-L}^{L}\sum_{b=-\ell}^{\ell} \text{SHT}[f]_L^M\,Y_L^a(n)\,\overline{Y_\ell^b(d_l)} \int_{SO(3)} D^{(L)}_{aM}(R)\,\overline{D^{(\ell)}_{b m}(R)}\,dR \nonumber\\
&= g_\ell\frac{4\pi}{2\ell+1} \cdot \sum_{L,M}\sum_{a=-L}^{L}\sum_{b=-\ell}^{\ell} \text{SHT}[f]_L^M\,Y_L^a(n)\,\overline{Y_\ell^b(d_l)} \frac{1}{2\ell+1}\,\delta_{L\ell}\,\delta_{a b}\,\delta_{M m} \nonumber\\
&= g_\ell \frac{4\pi}{{(2\ell+1)}^2} \cdot \sum_{a=-\ell}^{\ell} \text{SHT}[f](\ell, m)\,Y_\ell^a(n)\,\overline{Y_\ell^a(d_l)} \nonumber\\
&= g_\ell \frac{4\pi}{{(2\ell+1)}^2} \cdot \text{SHT}[f](\ell, m) \cdot\frac{2\ell+1}{4\pi}\,P_\ell(n\cdot d_l) \nonumber\\
&= \frac{g_\ell \cdot \text{SHT}[f](\ell, m)}{2\ell+1}\,P_\ell(n\cdot d_l)
\end{align}

This final derived solution exhibits separation of variables: the dependence on the input \(f\) appears only through its spherical harmonic coefficient \(f_\ell^m\), while the geometric dependence on the two fixed directions \(n\) and \(d\) is captured entirely by \(P_\ell(n\cdot d)\). Besides, the coefficient $g_\ell$ is also a symmetrical kernel. Therefore, in our fusing operator \ourmodel, the anisotropic operator solution is simplified to the core anisotropic component:
\begin{align}
\label{Solution_ani}
\text{SHT}[g(u)](l, m) = \text{SHT}[f](\ell, m)P_\ell(n\cdot \mathbf{d_\theta}).
\end{align}
where $\mathbf{d_\theta}$ is the learnable orientations. The anisotropic operators respect underlying geometric structure and allow the model to express distinct modes along distinct directions, reflecting the organization of biomedical substrates such as fiber tracts or cortical boundaries. Therefore, by explicitly modelling anisotropy where it is biomedically warranted, the operators enable the capture of heterogeneous, structure-driven variation. They thus complement spherical operators by restoring expressivity in a principled and biomedically grounded manner.


\newpage
\backmatter

\bibliography{sn-bibliography}

\end{document}

%% file: Tables/mnist_equ.tex
\newcommand{\tightbox}[1]{\begingroup\setlength{\fboxsep}{1pt}\colorbox{gray!20}{#1}\endgroup}

\begin{table*}[t]
\centering
    \caption{ACC↑ (in \%) on spherical mnist at different resolutions.
    }
    \label{tab:ACC}
    \setlength{\tabcolsep}{7pt}
    \def\arraystretch{1.2}
    
    \begin{tabular}{l!{\vrule}c!{\vrule}cccc}
\toprule
    
\multirow{2}{*}{\textbf{Method}} & \multirow{2}{*}{\textbf{Channel}}
& \multicolumn{4}{c}{\textbf{Different Resolutions}} \\
\cmidrule(lr){3-6}
& & 32×32 & 64×64 & 128×128 & 256×256  \\
    \cmidrule(lr){1-6}
         CNN       & 256 
        & 98.1 
        & 98.1 
        & 98.3 
        & 98.2
        \\
        
     Spherical CNN       & 64 
    & 95.1 
    & 95.3 
    & 95.6 
    & 95.8 
    \\

    SFNO    & 256 
    & 97.3 
    & 97.4 
    & 98.9
    & 99.1 
    \\
    
    \ournetwork~(Ours)    & 256
    & 99.3 
    & 99.4  
    & 99.6 
    & 99.7 
    \\
   
    \bottomrule
    \end{tabular}%
\end{table*}

%% file: Tables/SWE_results.tex
\begin{table*}[t]
\centering
\caption{MRE↓ ($\times 10^{-3}$) on SSWE at different spatial resolutions and prediction horizons. Bold indicates the best performance.}
\label{tab:SWE_MRE_all}
\resizebox{\textwidth}{!}{%
\begin{tabular}{l!{\vrule}cccc!{\vrule}cccc!{\vrule}cccc!{\vrule}cccc}
\toprule
\multirow{2}{*}{\textbf{Method}} 
& \multicolumn{16}{c}{\textbf{Prediction Horizon}} \\
\cmidrule(lr){2-17}
& \multicolumn{4}{c!{\vrule}}{\textbf{5 h}} 
& \multicolumn{4}{c!{\vrule}}{\textbf{10 h}} 
& \multicolumn{4}{c!{\vrule}}{\textbf{15 h}} 
& \multicolumn{4}{c}{\textbf{20 h}} \\
\cmidrule(lr){2-5} \cmidrule(lr){6-9} \cmidrule(lr){10-13} \cmidrule(lr){14-17}
& 32$\times$64 & 64$\times$128 & 128$\times$256 & 256$\times$512
& 32$\times$64 & 64$\times$128 & 128$\times$256 & 256$\times$512
& 32$\times$64 & 64$\times$128 & 128$\times$256 & 256$\times$512
& 32$\times$64 & 64$\times$128 & 128$\times$256 & 256$\times$512 \\
\midrule
FNO
& 0.72 & 0.82 & 1.97 & 2.59
& 1.00 & 1.30 & 2.24 & 3.36
& 1.55 & 1.63 & 2.71 & 4.47
& 1.59 & 1.82 & 3.04 & 5.14 \\

SFNO
& 0.59 & 0.55 & 0.68 & 0.65
& 0.74 & 0.70 & 0.79 & 0.72
& 1.04 & 0.87 & 0.88 & 0.86
& 1.37 & 1.12 & 1.00 & 0.98 \\

\midrule
\ourmodel
& \textbf{0.47} & \textbf{0.50} & \textbf{0.62} & \textbf{0.60}
& \textbf{0.59} & \textbf{0.59} & \textbf{0.67} & \textbf{0.64}
& \textbf{0.78} & \textbf{0.62} & \textbf{0.69} & \textbf{0.67}
& \textbf{0.99} & \textbf{0.66} & \textbf{0.77} & \textbf{0.86} \\

\bottomrule
\end{tabular}%
}
\end{table*}

%% file: Tables/FOD.tex
\begin{table*}[t]
\centering
\caption{Performance comparison on HCP dataset for brain nerve fiber modeling (ACC$\uparrow$). Bold represents the best result.}
\label{table_dmri}
\resizebox{\textwidth}{!}{%
\begin{tabular}{lccccc}
    \toprule
    Methods & \makecell{SSMT-CSD~\cite{khan2020three}}
            & \makecell{FOD-Net~\cite{zeng2022fod}}
            & \makecell{ESCNN~\cite{snoussi2025equivariant}}
            & \makecell{\bfseries \ourmodel}  \\
    \midrule
        Parameters    & --  & 19.44 M & 1.47 M & 1.11 M \\
    \midrule
    White matter 
    &{0.7523 $\pm$ 0.0256}  
    &{0.8858 $\pm$ 0.0138}
    &{0.9006 $\pm$ 0.0142} 
    & {\bfseries 0.9176 $\pm$ 0.0134}\\
    Whole brain    
    &{0.6640 $\pm$ 0.0145}  
    &{0.8250 $\pm$ 0.0159}
    &{0.8362 $\pm$ 0.0162} 
    &{\bfseries 0.8590 $\pm$ 0.0123} \\
    \bottomrule
    \end{tabular}%
}
\end{table*}

%% file: Tables/CC_mind101.tex
\begin{table*}[t]
\centering
\caption{Performance comparison on Mindboggle-101 dataset for brain cortical parcellation (ACC and Dice $\uparrow$, in \%). Bold represents the best result. (average of 1/3/5/7/9 in 9 folds)}
\label{table_dmri}
\setlength{\tabcolsep}{2pt}
\scriptsize
\resizebox{\textwidth}{!}{%
    \begin{tabular}{lcccccccc}
    \toprule
    Methods & \makecell{SPHARM-Net-c64~\cite{ha2022spharm}}
            & \makecell{SPHARM-Net-c128~\cite{ha2022spharm}}
            & \makecell{Spherical U-Net~\cite{zhao2019spherical}}
            & \makecell{\ourmodel-c64~(Ours)}
            & \makecell{\ourmodel-c128~(Ours)} \\ 
    \midrule
        Parameters    & 1.10 M & 4.31 M & 1.67 M & 1.03 M & 2.85 M \\
    \midrule
    ACC  & 89.55  & 89.88 & 89.42 & 89.94 & 90.42 \\
    \midrule
    Dice  & 88.42  & 88.49 & 88.36 & 88.61 & 88.87 \\
    
    \bottomrule
    \end{tabular}%
}
\end{table*}

%% file: Tables/CC_NAMIC39.tex
\begin{table*}[t]
\centering
\caption{Performance comparison on NAMIC-39 dataset for brain cortical parcellation (ACC and Dice $\uparrow$, in \%). Bold represents the best result. (average of 1/3/5/7/9 in 9 folds)}
\label{table_CC_NAMIC}
\setlength{\tabcolsep}{4pt}
\scriptsize
\resizebox{\textwidth}{!}{%
    \begin{tabular}{lcccccc}
    \toprule
    Methods & \makecell{SPHARM-Net-c64~\cite{ha2022spharm}}
            & \makecell{SPHARM-Net-c128~\cite{ha2022spharm}}
            & \makecell{Spherical U-Net~\cite{zhao2019spherical}}
            & \makecell{\ourmodel-c64~(Ours)}
            & \makecell{\ourmodel-c128~(Ours)} \\ 
    \midrule
        Parameters    & 1.10 M & 4.32 M & 1.67 M & 1.03 M & 2.85 M\\
    \midrule
    ACC  & 89.65  & 89.79 & 89.42 &  89.77 & 90.28 \\
    \midrule
    Dice  & 85.31  & 85.36 & 84.66 &  85.40 & 85.82 \\
    
    \bottomrule
    \end{tabular}%
}
\end{table*}

%% file: Tables/molecule.tex
\begin{table*}[t]
\centering
\caption{Performance comparison on QM7 for molecule feature analysis (RMSE$\downarrow$). Bold represents the best result.}
\label{table_qm7}
\setlength{\tabcolsep}{4pt}
\scriptsize
    \begin{tabular}{lcccc}
    \toprule
    Methods & \makecell{mlp}
            & \makecell{Spherical CNN}
            & \makecell{\bfseries \ourmodel         \bfseries (Ours)}
            \\       
    \midrule
        Parameters   
        & 0.001 M & 0.169 M 
        & 0.119 M  \\
    \midrule
    Average & 16.06 & 8.47 
    & 3.51  \\
    
    \bottomrule
    \end{tabular}%

\end{table*}